\documentclass[runningheads]{llncs}
\usepackage[T1]{fontenc}
\usepackage{boldline}

\usepackage[binary-units=true, detect-all]{siunitx}[=v2]

\usepackage{amsfonts}
\usepackage{amssymb}
\usepackage{amsmath}
\usepackage{graphicx}
\usepackage{cite} 
\begin{document}
\title{Live image-based neurosurgical guidance and roadmap generation using unsupervised embedding\thanks{Partially supported by the EANS 2021 Leica Research Grant}}
%
\titlerunning{Neurosurgical Roadmap Generation}
%
\author{Gary Sarwin\inst{1} \and
Alessandro Carretta\inst{2,3} \and
Victor Staartjes\inst{2}\and
Matteo Zoli\inst{3}\and
Diego Mazzatenta\inst{3}\and
Luca Regli\inst{2}\and
Carlo Serra\inst{2} \and 
Ender Konukoglu\inst{1}}
\authorrunning{G Sarwin et al.}
%
\institute{Computer Vision Lab, ETH Zurich, Switzerland \and
Department of Neurosurgery, University Hospital of Zurich, Zurich, Switzerland \and 
Department of Biomedical and Neuromotor Sciences (DIBINEM), University of Bologna, Bologna, Italy}

\maketitle              
\begin{abstract}
Advanced minimally invasive neurosurgery navigation relies mainly on Magnetic Resonance Imaging (MRI) guidance. 
MRI guidance, however, only provides pre-operative information in the majority of the cases.
Once the surgery begins, the value of this guidance diminishes to some extent because of the anatomical changes due to surgery. 
Guidance with live image feedback coming directly from the surgical device, e.g., endoscope, can complement MRI-based navigation or be an alternative if MRI guidance is not feasible. 
With this motivation, we present a method for live image-only guidance leveraging a large data set of annotated neurosurgical videos.
First, we report the performance of a deep learning-based object detection method, YOLO, on detecting anatomical structures in neurosurgical images. 
Second, we present a method for generating \emph{neurosurgical roadmaps} using unsupervised embedding without assuming exact anatomical matches between patients, presence of an extensive anatomical atlas, or the need for simultaneous localization and mapping.
A generated roadmap encodes the \emph{common} anatomical paths taken in surgeries in the training set. 
At inference, the roadmap can be used to map a surgeon's current location using live image feedback on the path to provide guidance by being able to predict which structures should appear going forward or backward, much like a mapping application. 
Even though the embedding is not supervised by position information, we show that it is correlated to the location inside the brain and on the surgical path. 
We trained and evaluated the proposed method with a data set of 166 transsphenoidal adenomectomy procedures.

\end{abstract}
\section{Introduction}
Specialists with extensive experience and a specific skill set are required to perform minimally invasive neurosurgeries. During these surgical procedures, differentiation between anatomical structures, orientation and localization is extremely challenging. 
On one side, excellent knowledge of the specific anatomy as visualized by the image feedback of the surgical device is required. On the other hand, low contrast, non-rigid deformations, a lack of clear boundaries between anatomical structures, and disruptions such as bleeding, make recognition even for experienced surgeons occasionally very challenging. Various techniques have been developed to help neurosurgeons become oriented and to perform surgery.
Computer-assisted neuronavigation has been an important tool and research topic for more than a decade \cite{Hartl2013WorldwideSurgery, Orringer2012NeuronavigationTrends
}, but it is still preoperative imaging-based, deeming it unreliable once the arachnoidal cisterns are opened and brain shift occurs \cite{Iversen2018AutomaticNeuronavigation}. 
More real-time anatomical guidance can be provided by intraoperative MRI \cite{Berkmann2014IntraoperativeAdenoma, Staartjes2021MachineSurgery, Stienen2019TheNote} and ultrasound \cite{Ulrich2012ResectionUltrasound, Burkhardt2014High-frequencyApproach}. 
Orientation has also been greatly enhanced by the application of fluorescent substances such as 5-aminolevulinic acid \cite{Stummer2017RandomizedGliomas,Hadjipanayis2015WhatGliomas}. 
Awake surgery \cite{Hervey-Jumper2015AwakePeriod} and electrophysiological neuromonitoring \cite{DeWittHamer2012ImpactMeta-analysis,Sanai2008FunctionalResection} can also help navigating around essential brain tissue.
These techniques work well and rely on physical traits, other than light reflection. 
However, they are expensive to implement, require the operating surgeon to become fluent in a new imaging modality, and may require temporarily halting the surgery or retracting surgical instruments to get the intra-operative information. \cite{Staartjes2020MachineSurvey}

Real-time anatomic recognition based on live image feedback from the surgical device has the potential to address these disadvantages and to act as a reliable tool for intraoperative orientation. 
This makes the application of machine vision algorithms appealing. 

The concepts of machine vision can likewise be employed in the neurosurgery operating room to analyze the digital image taken by the micro- or endoscope for automatically identifying the visible anatomic structures and mapping oneself on a planned surgical path. \cite{Staartjes2020MachineSurvey}

Deep learning applications within the operating room have become more prevalent in recent years. The applications include instrument and robotic tool detection and segmentation \cite{Wang2017DeepVideos, Sarikaya2017DetectionDetection}, surgical skill assessment \cite{Fawaz2018EvaluatingNetworks}, surgical task analysis  \cite{Luongo2021DeepSurgery}, and procedure automation \cite{ThananjeyanMultilateralTensioning}. Instrument or robotic tool detection and segmentation have been extensively researched for endoscopic procedures owing to the availability of various datasets and challenges \cite{Rivas-Blanco2021ASurgery}. Despite this research on endoscopic videos, the task of anatomic structure detection or segmentation, which could be a foundation for a new approach to neuronavigation, remains relatively unexplored and continues to be a challenge. 
Note that, anatomy recognition in surgical videos is significantly more challenging than the task of surgical tool detection because of the lack of clear boundaries and differences in color or texture between anatomical structures. 

The desire to provide a cheaper real-time solution without relying on additional machines and the improvement of deep learning techniques has driven also the development of vision-based localization methods. 
Approaches include structure from motion \cite{Leonard2018EvaluationData} and SLAM \cite{Grasa2011EKFSequences}, such as \cite{Ozyoruk2021EndoSLAMVideos, Mahmoud2016ORBSLAM-basedReconstruction}, for 3D map reconstruction based on feature correspondence. 
Many vision-based localization methods rely on landmarks or the challenging task of depth and pose estimation. 
The main idea behind these methods is to find distinctive landmark positions and follow them across frames for localization, which negatively impacts their performance owing to the low texture, a lack of distinguishable features, non-rigid deformations, and disruptions in endoscopic videos \cite{Ozyoruk2021EndoSLAMVideos}. 
These methods have mostly been applied to diagnostic procedures, such as colonoscopy, instead of surgical procedures, which pose significant difficulties. 
Abrupt changes due to the surgical procedure, e.g., bleeding and removal of tissue, make tracking landmarks extremely challenging or even impossible. 
Therefore, an alternative solution is required to address these challenges. 

In this study, a live image-only deep learning approach is proposed to provide guidance during endoscopic neurosurgical procedures. 
This approach relies on the detection of anatomical structures from RGB images in the form of bounding boxes instead of arbitrary landmarks, as was done in other approaches \cite{Grasa2014VisualEndoscopeb, Ozyoruk2021EndoSLAMVideos}, which are difficult to identify and track in the abruptly changing environment during a surgery. 
The bounding box detections are then used to map a sequence of video frames onto a 1-dimensional trajectory, that represents the surgical path. 
This allows for localization along the surgical path, and therefore predict anatomical structures in forward or backward directions. 
The surgical path is learned in an unsupervised manner using an autoencoder architecture from a training set of videos. 
Therefore, instead of reconstructing a 3D environment and localizing based on landmarks, we rely on a common surgical roadmap and localize ourselves within that map using bounding box detections.

\begin{figure}[!htb]
    \centering
    \includegraphics[width=\textwidth]{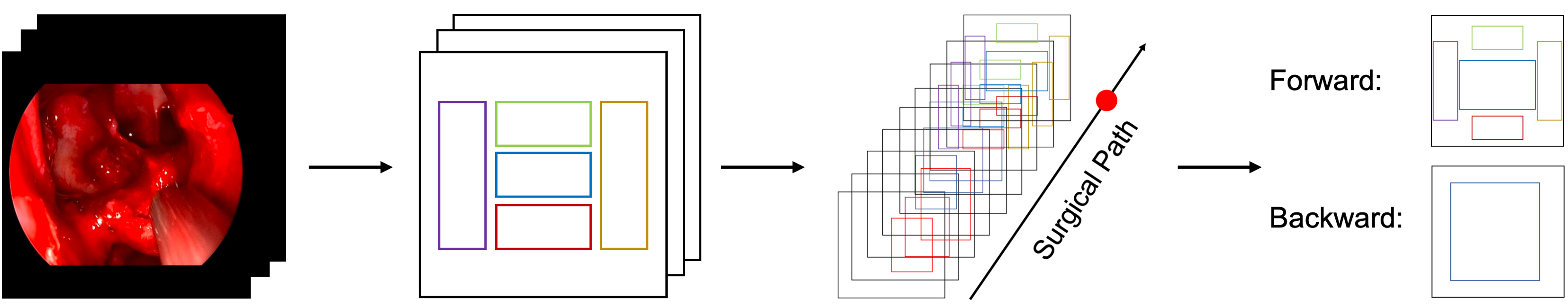}
    \caption{Simplified representation of the suggested approach. 1. A sequence of input images is processed to detect bounding boxes of anatomical structures. 2. A neural network encodes the sequence of detections into a latent variable that correlates with the position along the surgical path. 3. Given the current position along the surgical path, an estimation of anatomical structures in the forward or backward directions can be obtained, by extrapolating the current value of the latent variable. }
    \label{fig:fig1}
\end{figure}

The learned mapping rests on the principle that the visible anatomy and their relative sizes are strongly correlated with the position along the surgical trajectory. 
Towards this end, bounding box detections capture the presence of structures, their sizes, also relative to each other, and constellations.
A simplified representation is shown in Fig. \ref{fig:fig1}. 
Using bounding box detection of anatomical structures as semantic features mitigates the problem of varying appearance across different patients since bounding box composition is less likely to change across patients than the appearance of the anatomy in RGB images.
Furthermore, because the considered anatomical structures only have one instance in every patient, we do not need to rely on tracking of arbitrary structures, e.g., areas with unique appearance compared to their surrounding, which further facilitates dealing with disruptions during surgery, such as bleeding or flushing. We applied the proposed approach on the transsphenoidal adenomectomy procedure, where the surgical path is relatively one-dimensional, as shown in Fig. \ref{fig:fig2}, which makes it well-suited for the proof-of-concept of the suggested method. 

\section{Methods}
\subsection{Problem Formulation and Approach}
\begin{figure}[!b]
    \centering
    \includegraphics[width=0.8\textwidth]{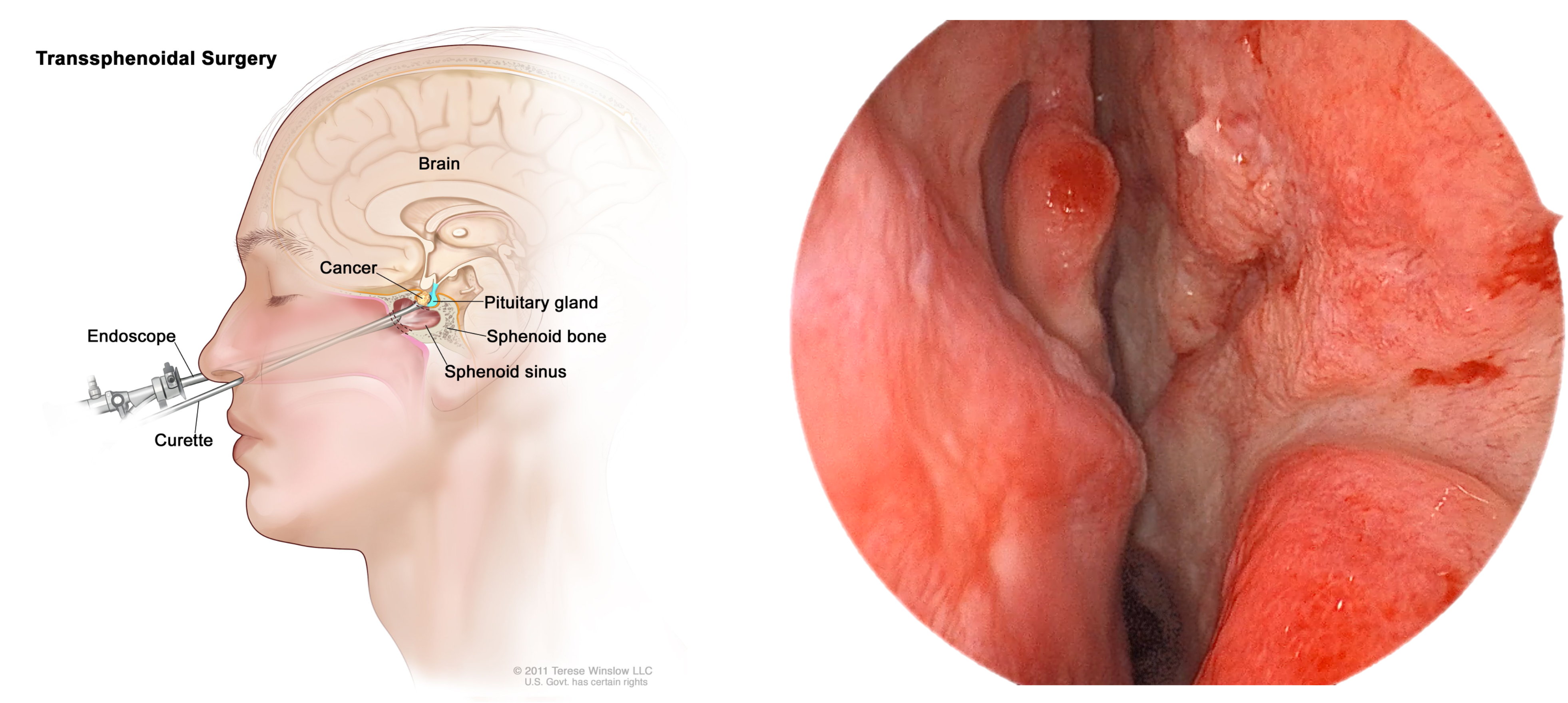}
    \caption{Left: Transsphenoidal adenomectomy procedure is performed to remove a tumor from the pituitary gland, located at the base of the brain. Through the use of an endoscope and various instruments, the surgeon inserts the instruments into the nostril and crosses the sphenoidal sinus to access the pituitary gland located behind the sella floor. Right: A video frame showing only the anatomy. Note that there is lack of clear differences between anatomical structures in such images.}
    \label{fig:fig2}
\end{figure}

Let $\mathbf{S}_{t}$ denote an image sequence that consists of endoscopic frames $\mathbf{x}_{t-s:t}$, such as the one shown in Figure~\ref{fig:fig2}, where $s$ represents the sequence length in terms of the number of frames, and $\mathbf{x}_t \in \mathbb{R}^{w \times h \times c}$ is the $t$-th frame with $w, h$, and $c$ denoting the width, height, and number of channels, respectively. 
Our main aim is to embed the sequence $\mathbf{S}_{t}$ in a 1D latent dimension represented by the variable $\mathbf{z}$.
This 1-D latent space represents the surgical path taken from the beginning of the procedure until the final desired anatomy is reached. 
The approach we take is to determine the anatomical structures visible in the sequence $\mathbf{S}_t$ along the surgical path and map the frame $\mathbf{x}_t$ to the latent space, where effectively the latent space acts as an \emph{implicit} anatomical atlas. 
We refer to this as an implicit atlas because the position information along the surgical path is not available for construction of the latent space. 
To achieve this, we perform object detection on all frames $\mathbf{x}_{t-s:t}$ in  $\mathbf{S}_{t}$ and obtain a sequence of detections $\mathbf{c}_{t-s:t}$ that we denote as $\mathbf{C}_t$. A detection $\mathbf{c}_t \in \mathbb{R}^{n \times 5}$ represents the anatomical structures and bounding boxes of the $t$-th frame, where $n$ denotes the number of different classes in the surgery. More specifically, $\mathbf{c}_t$ consists of a binary variable $\mathbf{y}_t = [y_t^0,\dots,y_t^n] \in \{0,1\}^{n}$ denoting the present structures (or classes) in the $t$-th frame and $\mathbf{b}_t = [\mathbf{b}_t^0,\dots,\mathbf{b}_t^n]^T \in \mathbb{R}^{n \times 4}$ denoting the respective bounding box coordinates. 
An autoencoder architecture was used to achieve the embedding, i.e., mapping $\mathbf{C}_t$ to $\mathbf{z}_t$. The encoder maps $\mathbf{C}_t$  to $\mathbf{z}_{t}$, and the decoder generates $\hat{\mathbf{c}}_t$, which represents the detections of the last frame in a given sequence, given $\mathbf{z}_{t}$. The model parameters are updated to ensure that $\hat{\mathbf{c}}_t$ fits $\mathbf{c}_{t}$ on a training set as will be explained in the following.

\subsection{Object Detection}
Our approach requires being able to detect anatomical structures as bounding boxes in frames from a video. To this end, the object detection part of the pipeline is fulfilled by an iteration of the YOLO network \cite{Redmon2015YouDetection}. Specifically, the YOLOv7 network was used \cite{Wang2022YOLOv7:Detectors}. The network was trained on the endoscopic videos in the training set, where frames are sparsely labeled with bounding boxes, which contains 15 different anatomical classes and one surgical instrument class. The trained network was then applied to all the frames of the training videos to create detections of these classes on every frame of the videos. These are then used to train the subsequent autoencoder that models the embedding.

\subsection{Embedding}
\begin{figure}[!b]
    \centering
    \includegraphics[width=0.8\textwidth]{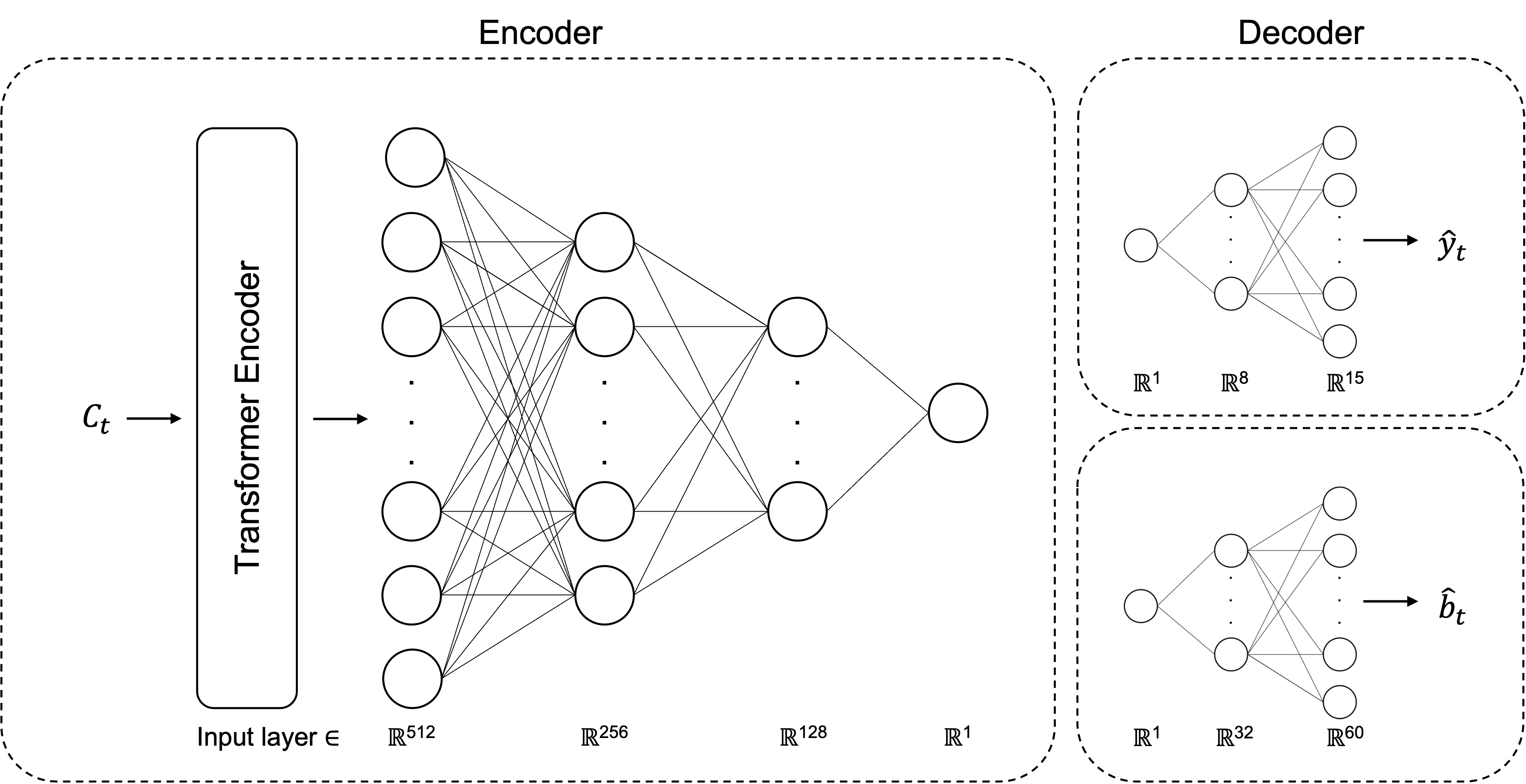}
    \caption{The model architecture. The model consists of an encoder and two decoders. The encoder consists of a multi-head attention layer, i.e., a transformer encoder, which takes $\mathbf{C}_t$ as input, followed by a series of fully connected layers to embed the input in a 1D latent dimension. The two decoders consist of fully connected layers to generate the class probabilities $\hat{\mathbf{y}}_t$ and the bounding box coordinates $\hat{\mathbf{b}}_t$, respectively.}
    \label{fig:fig4}
\end{figure}

To encode the bounding boxes onto the 1D latent space, the output from the YOLO network was slightly modified to exclude the surgical instrument, because the presence of the instrument in the frame is not necessarily correlated with the frame's position along the surgical path. 
The autoencoder was designed to reconstruct only the last frame $\mathbf{c}_{t}$ in $\mathbf{C}_t$ because $\mathbf{z}_{t}$ is desired to correspond to the current position. However, it takes into account $s$ previous frames to provide more information while determining the latent representation $\mathbf{z}_t$ of an $\mathbf{x}_t$. 

The encoder of the autoencoder network consists of multi-head attention layers followed by fully connected layers, which eventually reduce the features to a 1D value. Here a transformer-based encoder is used to encode the temporal information in the sequence of detections. The decoder consists of two fully connected decoders, the first of which generates the class probabilities $\hat{\mathbf{y}}_t$ of $\hat{\mathbf{c}}_t$ and the second generates the corresponding bounding boxes $\hat{\mathbf{b}}_t$.  A simplified representation of the network is shown in Fig. \ref{fig:fig4}.  The loss function consists of a classification loss and a bounding box loss, which is only calculated for the classes present in the ground truth. This results in the following objective to minimize for the $t$-th frame in the $m$-th training video:
\begin{equation*}
\mathcal{L}_{m,t}=-\sum_{i=1}^n\left(y_{m,t}^i \log \left(\hat{y}_{m,t}^i\right)+\left(1-y_{m,t}^i\right) \log \left(1-\hat{y}_{m,t}^i\right)\right)+\sum_{i=1}^n y_{m,t}^i\left|\mathbf{b}_{m,t}^i-\hat{\mathbf{b}}_{m,t}^i\right|,
\end{equation*}
\noindent where $|\cdot|$ is the $l_1$ loss and $\hat{y}^i_{m,t}$ and $\hat{\mathbf{n}}^i_{m,t}$ are generated from $\mathbf{z}_{m,t}$ using the autoencoder. The total training loss is then obtained by summing $\mathcal{L}_{m,t}$ over all frames and training videos. The proposed loss function can be considered to correspond to maximizing the joint likelihood of a given $\mathbf{y}$ and $\mathbf{b}$ with a probabilistic model that uses a mixture model for the bounding boxes.

\section{Experiments and Results}

\subsection{Dataset}

The object detection dataset used consists of 166 anonymized videos recorded during a transsphenoidal adenomectomy in 166 patients. The videos were recorded using various endoscopes and at multiple facilities, and made available through general research consent. The videos were labeled by neurosurgeons and include 16 different classes, that is, 15 different anatomical structure classes and one surgical instrument class. In total the dataset consists of approximately 19000 labeled frames, and around 3$\times 10^6$ frames in total. All the classes have only one instance in every video because of the anatomical nature of the human body, except for the instrument class, because of the various instruments being used during the procedures. Out of the 166 videos,  146 were used for training and validation, and 20 for testing. While we used different centers in our data, we acknowledge that all the centers are concentrated in one geographic location, which may induce biases in our algorithms. However, we also note that we use different endoscopes and they were acquired throughout the last 10 years. 

\subsection{Implementation Details}

The implementation of the YOLO network follows \cite{Wang2022YOLOv7:Detectors} using an input resolution of 1280x1280. The model reached convergence after 125 epochs. To generate the data to train the autoencoder, the object confidence score and intersection-over-union (IoU) threshold were set to 0.25 and 0.45, respectively.

The autoencoder uses a transformer encoder that consists of six transformer encoder layers with five heads and an input size of $s \times 15 \times 5$, where $s$ is set to 64 frames. Subsequently,  the dimension of the output of the transformer encoder is reduced by three fully connected layers to 512, 256, and 128 using rectified linear unit (ReLU) activation functions in between. Finally, the last fully connected layer reduces the dimension to 1D and uses a sigmoid activation function to obtain the final latent variable. Furthermore, the two decoders, the class decoder and bounding box decoder, consist of two fully connected layers, increasing the dimension of the latent variable from 1D to 8, 15, and 32, $15 \times 4$, respectively. The first layer of both decoders is followed by a ReLU activation function and the final layer by a sigmoid activation function. 

For training of the autoencoder, the AdamW optimizer \cite{Loshchilov2017DecoupledRegularization} was used in combination with a warm-up scheduler that linearly increases the learning rate from 0 to \num{1e-4} over 60 epochs. The model was trained for 170 epochs.

\subsection{Results}
\paragraph{Anatomical Structure Detection: }
The performance of the YOLO network on the test videos is shown in Table \ref{tab1}, using an IoU threshold for non-maximum suppression of 0.45 and an object confidence threshold of 0.001. The latter is set to 0.001 as this is the common threshold used in other detection works. It is surprising how well YOLO model works on the challenging problem of detecting anatomical structures in endoscopic neurosurgical videos. 
\begin{table}[!b]
\begin{center}
\caption{YOLO detection model results on 20 test videos with an IoU threshold for non-maximum suppression of 0.45 and an object confidence threshold of 0.001.}\label{tab1}
\begin{tabular}{|lll V{4} lll|}
\hline
Class &  $AP_{50}$ & $AP_{50:95}$ & Class & $AP_{50}$ & $AP_{50:95}$\\
\hlineB{4}
All (mean)&  53.4 & 26.2 & Ostium  & 43.1 & 19.4\\
Septum & 78.6 & 57.9&Instrument & 94.4 &  55.4\\
SupM & 40.6 & 21.7&Rostrum & 17.9 & 4.70\\
MidM & 63.8 & 36.7&Sphenoidal Sinus & 74.0 & 40.9\\
InfM & 62.9 & 33.7&Sella Floor & 70.6 & 27.6\\
Coana & 54.3 & 22.2&Clival Recess & 58.9 & 23.4\\
Floor & 65.3 & 31.2&Planum & 34.9 & 15.1\\
RecSphEthm & 41.1 & 14.9&Osseous Carotis Right & 21.6 & 5.37\\
Osseous Carotis Left & 32.4 & 9.63& &  & \\
\hline
\end{tabular}
\end{center}
\end{table}

\begin{figure}[h!]
    \centering
    \includegraphics[width=1\textwidth]{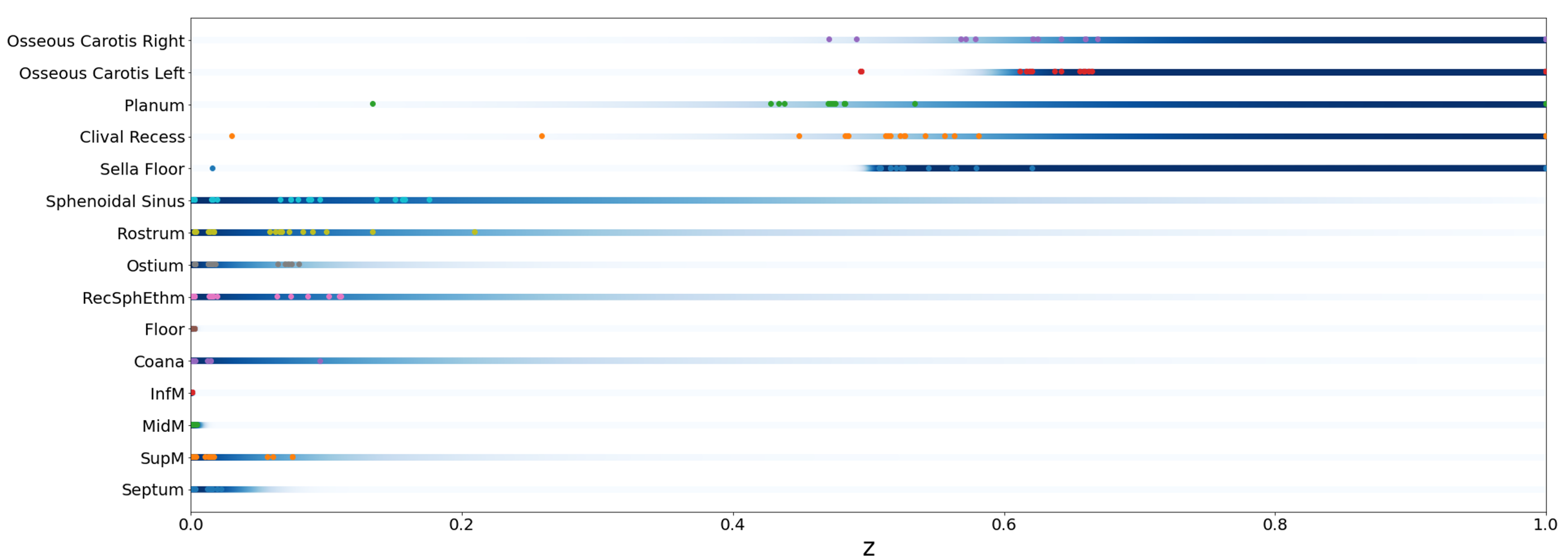}
    \caption{The normalized generated confidences of each class along the latent space. This visualizes the probability of finding a certain anatomical structure at a specific point in the latent space. Additionally, video frames of twenty test videos responsible for the first appearances of anatomical structures in every video have been encoded and overlaid onto the confidence intervals to demonstrate that their locations correlate with the beginning of these intervals. }
    \label{fig:fig5}
\end{figure}

\paragraph{Qualitative Assessment of the Embedding: }
First, to evaluate the learned latent representation, we compute the confidences for every class, i.e., $y^i$, for different points on the latent space, and plot them in Fig. \ref{fig:fig5}. The confidences are normalized for every class, where the maximum confidence of a class corresponds to the darkest shade of blue, and vice versa. This shows how likely it is to find an anatomical structure at a certain location in the latent space, resembling a confidence interval for the structure's presence along the surgical path. 

Fig. \ref{fig:fig5} shows how the autoencoder encodes and separates anatomical structures along the surgical path. For example, from left ($z=0$, the start of the surgical path) to right ($z=1$, the end of the surgical path), it can be seen that the septum is frequently visible at the start of the surgical path, but later it is no longer be visible. Because a sequence encodes to a single point in the latent space, positioning along the surgical path is possible and allows the forecasting of structures in both the forward and backward directions.

Furthermore, twenty test videos were used to validate the spatial embedding of the anatomical structures. For every single one of the videos, the frame of the first appearance of every anatomical structure was noted. To obtain the corresponding $z$-value for each of the noted frames, a sequence was created from the same frame, using the $s$ previous frames, for all of the noted frames. These sequences were then embedded into the 1D latent dimension to determine whether their locations corresponded to the beginning of the confidence intervals in the latent space where corresponding anatomical structures were expected to start appearing. When examining the encodings of the video data, it is evident that the points are located on the left side of the confidence interval for every class. When considering a path from $z=0$ to $z=1$, this demonstrates that the autoencoder is able to accurately map the first appearances of every class in the validation videos to the beginning of the confidence intervals for the classes, showing the network is capable of relative positional embedding.

\begin{figure}[!b]
    \centering
    \includegraphics[width=1\textwidth]{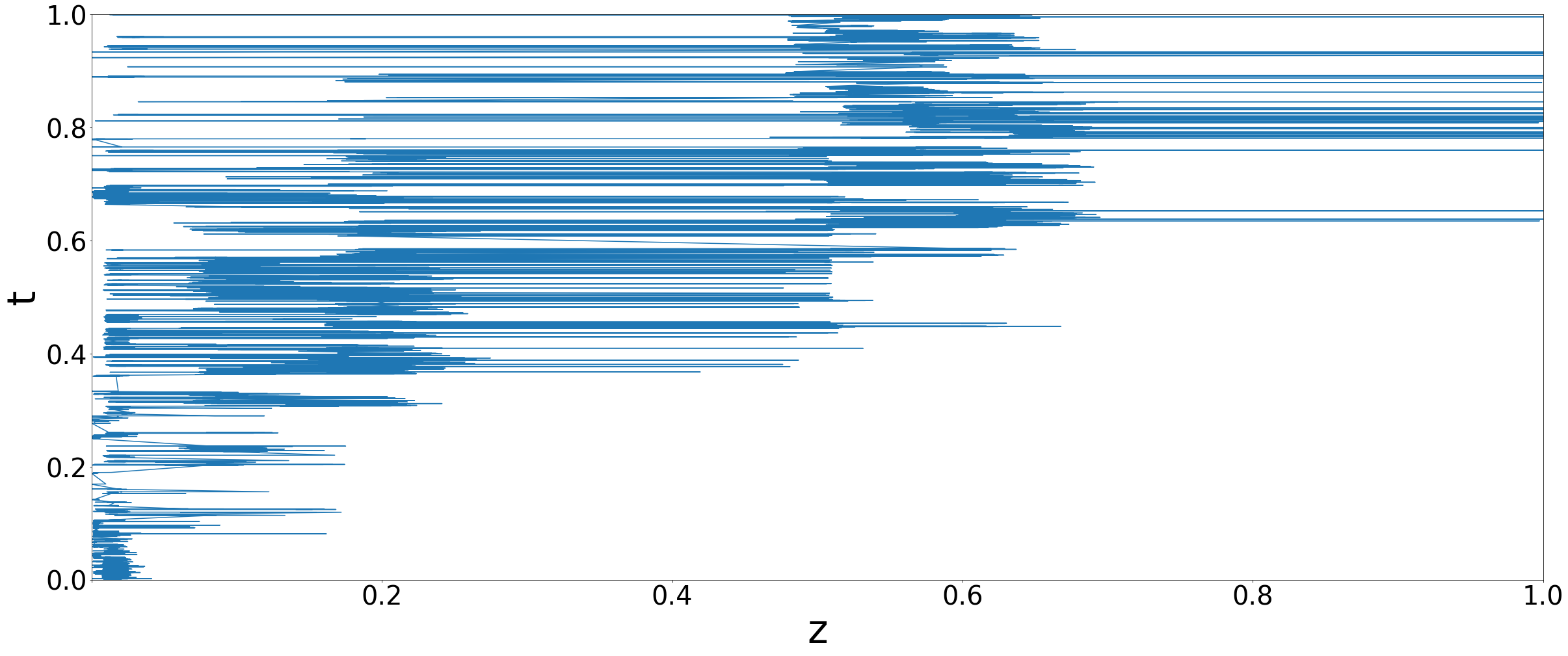}
    \caption{Z-values over time during a surgical video. Certain $z$-values are encoded more frequently than others, such as approximately $z=0.2$ and $z=0.6$, which is related to the amount of time spent at a certain location during the surgery.}
    \label{fig:fig11}
\end{figure}

Fig. \ref{fig:fig11} plots the $z$-value against time (t) over an entire surgical video, where $t=1$ denotes the end of the video. The plot on the right shows how the endoscope is frequently extracted going from a certain z-value back to 0, the beginning of the surgical path, instantaneously. Retraction of the endoscope and replacement is common in this surgery and the plot reflects this. Subsequently, swift reinsertion of the endoscope to the region of interest is performed, spending little time at $z$-values inferior to the ones visited before extraction. Additionally, locations along the latent space are visible where more time is spent than others, such as around $z=0.2$ and around $z=0.6$, which correspond to locations where tissue is removed or passageways are created, such as the opening of the sphenoidal sinus. We also note that the z-value also shoots to z=1 at certain times. Z-values from 0.5 to 1.0 actually correspond to a narrow section of the surgical path. However, this narrow section is the crux of the surgery where the surgeon spends more time. Hence the behavior of the model is expected since more time spent in this section leads to a higher number of images, and ultimately, covers a larger section of the latent space. 

Lastly, we used the decoder to generate bounding boxes moving along the 1D latent dimension from one end to the other. A GIF showing the bounding boxes of the classes that the model expected to encounter along the surgical path can be found here at the following link: \url{https://gifyu.com/image/SkMIX}. Certain classes are expected at different points, and their locations and sizes vary along the latent space. From appearance to disappearance of classes, their bounding boxes grow larger and their centers either move away from the center of the frame to the outer region of the frame or stay centered. This behavior is expected when moving through a tunnel structure, as in an endoscopic procedure. This shows that the latent space can learn a roadmap of the surgery with the expected anatomical structures at any location on the surgical path.

\begin{figure}[!t]
    \centering
    \includegraphics[width=\textwidth]{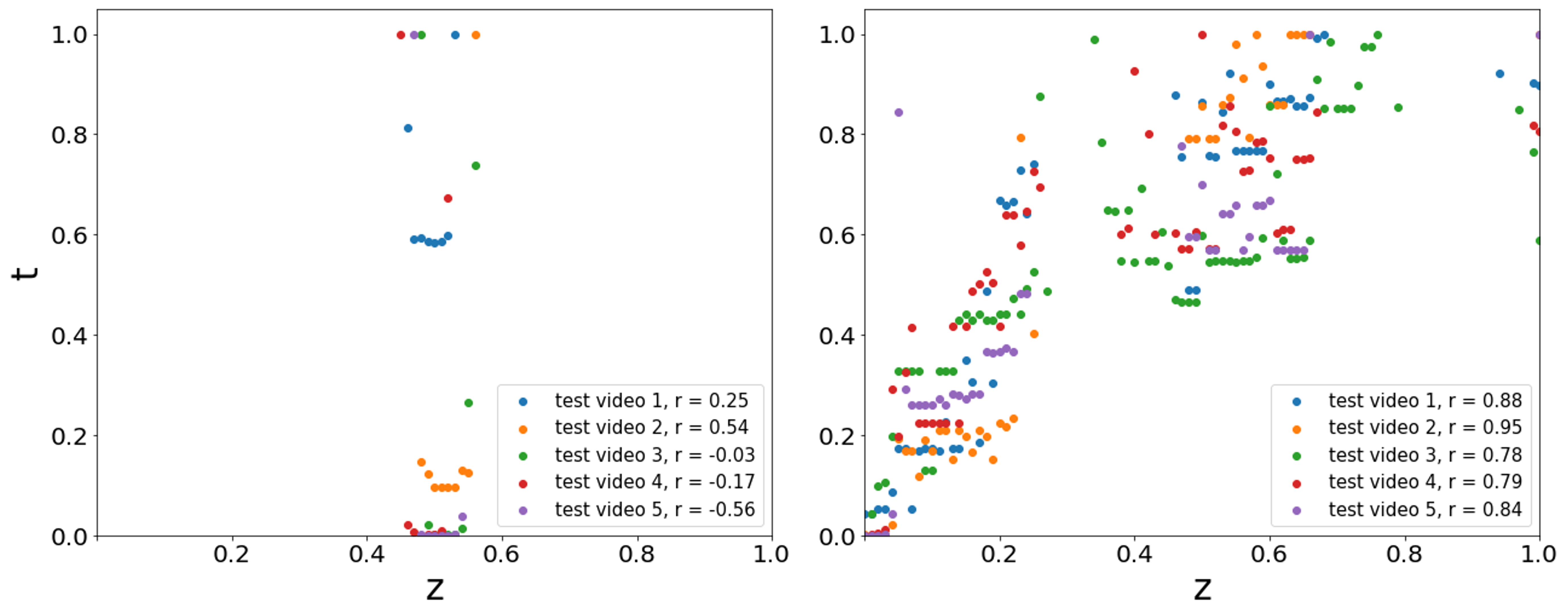}
    \caption{Latent variable plotted against time of its first encoding for 5 surgical videos. Pearson correlation coefficients for first-time of appearance and $\mathbf{z}$ values are given for an untrained (left) and trained (right) model. In these plots, $t=1$ denotes the end of a video. The untrained model provides a baseline for expected correlation coefficients. High correlation coefficients suggest the embedding captures the relative position on the surgical path.}
    \label{fig:fig7}
\end{figure}

\paragraph{Quantitative Assessment of the Embedding:}
Beyond qualitative analyses, we performed quantitative analysis to demonstrate that the latent space spatially embeds the surgery path. As there are no ground truth labels on the position of a video frame on the surgical path, a direct quantitative analysis comparing z-value to ground truth position on the surgical path is not possible. To provide a quantitative evaluation, we make the observation that if the latent space represents the surgical path spatially, frames encoding $z$-values at the beginning of the path should be encountered in the early stages of the surgery, and vice versa. Therefore, the timestamp $t$ of a sequence responsible for the first encoding of a specific $z$-value should increase with increasing $z$-value. This is confirmed by the mean correlation coefficient between $t$ and $\mathbf{z}$ for the 20 videos, which is $0.80$. \newline Fig. \ref{fig:fig7} shows the relation between $t$ and $\mathbf{z}$ for five test videos with their corresponding Pearson correlation coefficients $r$ for an untrained and a trained model.

\section{Conclusion}
In this study, we propose a novel approach to neuronavigation based on deep learning. The suggested approach is live image-based and uses bounding box detections of anatomical structures to localize itself on a common surgical roadmap that is learned from a dataset containing numerous videos from a specific surgical procedure. The mapping is modeled by the use of an autoencoder architecture and trained without supervision. The method allows for the localization and forecasting of anatomical structures that are to be encountered in forward and backward directions along the surgical path, similar to a mapping application.

The presented work has also some limitations. The main limitation is that we focused on only one surgery in this initial work. Extension to other surgeries is our future research topic. The proposed method can also be combined with SLAM approaches as well as guidance provided by MRI. Both of these directions also form our future work. Another limitation is that the latent dimension only provides relative positional encoding. Going beyond this may require further labels on the real position on the surgical path.


%
%
%
\bibliographystyle{splncs04}

\begin{thebibliography}{10}
\providecommand{\url}[1]{\texttt{#1}}
\providecommand{\urlprefix}{URL }
\providecommand{\doi}[1]{https://doi.org/#1}

\bibitem{Berkmann2014IntraoperativeAdenoma}
Berkmann, S., Schlaffer, S., Nimsky, C., Fahlbusch, R., Buchfelder, M.:
  {Intraoperative high-field MRI for transsphenoidal reoperations of
  nonfunctioning pituitary adenoma}. Journal of neurosurgery  \textbf{121}(5),
  1166--1175 (11 2014)

\bibitem{Burkhardt2014High-frequencyApproach}
Burkhardt, J.K., Serra, C., Neidert, M.C., Woernle, C.M., Fierstra, J., Regli,
  L., Bozinov, O.: {High-frequency intra-operative ultrasound-guided surgery of
  superficial intra-cerebral lesions via a single-burr-hole approach}.
  Ultrasound in medicine {\&} biology  \textbf{40}(7),  1469--1475 (2014)

\bibitem{DeWittHamer2012ImpactMeta-analysis}
De~Witt~Hamer, P.C., Robles, S.G., Zwinderman, A.H., Duffau, H., Berger, M.S.:
  {Impact of intraoperative stimulation brain mapping on glioma surgery
  outcome: a meta-analysis}. Journal of clinical oncology : official journal of
  the American Society of Clinical Oncology  \textbf{30}(20),  2559--2565 (7
  2012)

\bibitem{Fawaz2018EvaluatingNetworks}
Fawaz, H.I., Forestier, G., Weber, J., Idoumghar, L., Muller, P.A.: {Evaluating
  surgical skills from kinematic data using convolutional neural networks}.
  Lecture Notes in Computer Science (including subseries Lecture Notes in
  Artificial Intelligence and Lecture Notes in Bioinformatics)  \textbf{11073
  LNCS},  214--221 (6 2018)

\bibitem{Grasa2014VisualEndoscopeb}
Grasa, O.G., Bernal, E., Casado, S., Gil, I., Montiel, J.M.: {Visual slam for
  handheld monocular endoscope}. IEEE Transactions on Medical Imaging
  \textbf{33}(1),  135--146 (1 2014)

\bibitem{Grasa2011EKFSequences}
Grasa, O.G., Civera, J., Montiel, J.M.: {EKF monocular SLAM with relocalization
  for laparoscopic sequences}. Proceedings - IEEE International Conference on
  Robotics and Automation pp. 4816--4821 (2011)

\bibitem{Hadjipanayis2015WhatGliomas}
Hadjipanayis, C.G., Widhalm, G., Stummer, W.: {What is the Surgical Benefit of
  Utilizing 5-Aminolevulinic Acid for Fluorescence-Guided Surgery of Malignant
  Gliomas?} Neurosurgery  \textbf{77}(5),  663--673 (8 2015)

\bibitem{Hartl2013WorldwideSurgery}
H{\"{a}}rtl, R., Lam, K.S., Wang, J., Korge, A., Kandziora, F., Audig{\'{e}},
  L.: {Worldwide survey on the use of navigation in spine surgery}. World
  neurosurgery  \textbf{79}(1),  162--172 (1 2013)

\bibitem{Hervey-Jumper2015AwakePeriod}
Hervey-Jumper, S.L., Li, J., Lau, D., Molinaro, A.M., Perry, D.W., Meng, L.,
  Berger, M.S.: {Awake craniotomy to maximize glioma resection: methods and
  technical nuances over a 27-year period}. Journal of neurosurgery
  \textbf{123}(2),  325--339 (8 2015)

\bibitem{Iversen2018AutomaticNeuronavigation}
Iversen, D.H., Wein, W., Lindseth, F., Unsg{\aa}rd, G., Reinertsen, I.:
  {Automatic Intraoperative Correction of Brain Shift for Accurate
  Neuronavigation}. World neurosurgery  \textbf{120},  e1071--e1078 (12 2018)

\bibitem{Leonard2018EvaluationData}
Leonard, S., Sinha, A., Reiter, A., Ishii, M., Gallia, G.L., Taylor, R.H.,
  Hager, G.D.: {Evaluation and Stability Analysis of Video-Based Navigation
  System for Functional Endoscopic Sinus Surgery on In Vivo Clinical Data}.
  IEEE Transactions on Medical Imaging  \textbf{37}(10),  2185--2195 (10 2018)

\bibitem{Loshchilov2017DecoupledRegularization}
Loshchilov, I., Hutter, F.: {Decoupled Weight Decay Regularization}. 7th
  International Conference on Learning Representations, ICLR 2019  (11 2017)

\bibitem{Luongo2021DeepSurgery}
Luongo, F., Hakim, R., Nguyen, J.H., Anandkumar, A., Hung, A.J.: {Deep
  learning-based computer vision to recognize and classify suturing gestures in
  robot-assisted surgery}. Surgery  \textbf{169}(5),  1240--1244 (5 2021)

\bibitem{Mahmoud2016ORBSLAM-basedReconstruction}
Mahmoud, N., Cirauqui, I., Hostettler, A., Doignon, C., Soler, L., Marescaux,
  J., Montiel, J.M.: {ORBSLAM-based Endoscope Tracking and 3D Reconstruction}.
  Lecture Notes in Computer Science (including subseries Lecture Notes in
  Artificial Intelligence and Lecture Notes in Bioinformatics)  \textbf{10170
  LNCS},  72--83 (8 2016)

\bibitem{Orringer2012NeuronavigationTrends}
Orringer, D.A., Golby, A., Jolesz, F.: {Neuronavigation in the surgical
  management of brain tumors: current and future trends}. Expert review of
  medical devices  \textbf{9}(5),  491--500 (9 2012)

\bibitem{Ozyoruk2021EndoSLAMVideos}
Ozyoruk, K.B., Gokceler, G.I., Bobrow, T.L., Coskun, G., Incetan, K.,
  Almalioglu, Y., Mahmood, F., Curto, E., Perdigoto, L., Oliveira, M., Sahin,
  H., Araujo, H., Alexandrino, H., Durr, N.J., Gilbert, H.B., Turan, M.:
  {EndoSLAM dataset and an unsupervised monocular visual odometry and depth
  estimation approach for endoscopic videos}. Medical Image Analysis
  \textbf{71},  102058 (7 2021)

\bibitem{Redmon2015YouDetection}
Redmon, J., Divvala, S., Girshick, R., Farhadi, A.: {You Only Look Once:
  Unified, Real-Time Object Detection}. Proceedings of the IEEE Computer
  Society Conference on Computer Vision and Pattern Recognition
  \textbf{2016-December},  779--788 (6 2015)

\bibitem{Rivas-Blanco2021ASurgery}
Rivas-Blanco, I., Perez-Del-Pulgar, C.J., Garcia-Morales, I., Munoz, V.F.,
  Rivas-Blanco, I.: {A Review on Deep Learning in Minimally Invasive Surgery}.
  IEEE Access  \textbf{9},  48658--48678 (2021)

\bibitem{Sanai2008FunctionalResection}
Sanai, N., Mirzadeh, Z., Berger, M.S.: {Functional outcome after language
  mapping for glioma resection}. The New England journal of medicine
  \textbf{358}(1),  18--27 (1 2008)

\bibitem{Sarikaya2017DetectionDetection}
Sarikaya, D., Corso, J.J., Guru, K.A.: {Detection and Localization of Robotic
  Tools in Robot-Assisted Surgery Videos Using Deep Neural Networks for Region
  Proposal and Detection}. IEEE Transactions on Medical Imaging
  \textbf{36}(7),  1542--1549 (7 2017)

\bibitem{Staartjes2020MachineSurvey}
Staartjes, V.E., Stumpo, V., Kernbach, J.M., Klukowska, A.M., Gadjradj, P.S.,
  Schr{\"{o}}der, M.L., Veeravagu, A., Stienen, M.N., van Niftrik, C.H., Serra,
  C., Regli, L.: {Machine learning in neurosurgery: a global survey}. Acta
  Neurochirurgica  \textbf{162}(12),  3081--3091 (12 2020)

\bibitem{Staartjes2021MachineSurgery}
Staartjes, V.E., Volokitin, A., Regli, L., Konukoglu, E., Serra, C.: {Machine
  Vision for Real-Time Intraoperative Anatomic Guidance: A Proof-of-Concept
  Study in Endoscopic Pituitary Surgery}. Operative neurosurgery (Hagerstown,
  Md.)  \textbf{21}(4),  242--247 (10 2021)

\bibitem{Stienen2019TheNote}
Stienen, M.N., Fierstra, J., Pangalu, A., Regli, L., Bozinov, O.: {The Zurich
  Checklist for Safety in the Intraoperative Magnetic Resonance Imaging Suite:
  Technical Note}. Operative neurosurgery (Hagerstown, Md.)  \textbf{16}(6),
  756--765 (6 2019)

\bibitem{Stummer2017RandomizedGliomas}
Stummer, W., Stepp, H., Wiestler, O.D., Pichlmeier, U.: {Randomized,
  Prospective Double-Blinded Study Comparing 3 Different Doses of
  5-Aminolevulinic Acid for Fluorescence-Guided Resections of Malignant
  Gliomas}. Neurosurgery  \textbf{81}(2),  230--239 (8 2017)

\bibitem{ThananjeyanMultilateralTensioning}
Thananjeyan, B., Garg, A., Krishnan, S., Chen, C., Miller, L., Goldberg, K.:
  {Multilateral Surgical Pattern Cutting in 2D Orthotropic Gauze with Deep
  Reinforcement Learning Policies for Tensioning}

\bibitem{Ulrich2012ResectionUltrasound}
Ulrich, N.H., Burkhardt, J.K., Serra, C., Bernays, R.L., Bozinov, O.:
  {Resection of pediatric intracerebral tumors with the aid of intraoperative
  real-time 3-D ultrasound}. Child's nervous system : ChNS : official journal
  of the International Society for Pediatric Neurosurgery  \textbf{28}(1),
  101--109 (1 2012)

\bibitem{Wang2022YOLOv7:Detectors}
Wang, C.Y., Bochkovskiy, A., Liao, H.Y.M.: {YOLOv7: Trainable bag-of-freebies
  sets new state-of-the-art for real-time object detectors}  (7 2022)

\bibitem{Wang2017DeepVideos}
Wang, S., Raju, A., Huang, J.: {Deep learning based multi-label classification
  for surgical tool presence detection in laparoscopic videos}. undefined pp.
  620--623 (6 2017)

\end{thebibliography}

\end{document}